\PassOptionsToPackage{dvipsnames}{xcolor}
\PassOptionsToPackage{shortlabels}{enumitem}
\documentclass[11pt, a4paper]{berkeley-paul}

\usepackage[ruled,vlined,algo2e]{algorithm2e}

\usepackage{mathtools}
\usepackage{bm}
\usepackage{cancel}
\usepackage{siunitx}

\usepackage[textsize=tiny]{todonotes}

\usepackage{blindtext}
\usepackage[labelformat=empty, position=top]{subcaption}
\usepackage[export]{adjustbox}

\usepackage{wrapfig}

\usepackage{nicefrac}       %

\usepackage{multirow}
\usepackage{multicol}
\usepackage{soul}

\usepackage{transparent}

\usepackage{tikz}
\usetikzlibrary{decorations.pathreplacing,calc}

\usepackage{hyperref}
\usepackage[capitalize,noabbrev]{cleveref}

\theoremstyle{plain}

\def\eqref#1{equation~\ref{#1}}

\def\1{\bm{1}}

\DeclareMathAlphabet{\mathsfit}{\encodingdefault}{\sfdefault}{m}{sl}
\SetMathAlphabet{\mathsfit}{bold}{\encodingdefault}{\sfdefault}{bx}{n}

\definecolor{darkblue}{rgb}{0,0.08,0.45}
\definecolor{cred}{HTML}{D62728}
\definecolor{cblue}{HTML}{1F77B4}
\definecolor{cgreen}{HTML}{79AB76}
\definecolor{cgrey}{rgb}{0.6,0.6,0.6}

\renewcommand{\mathbf}{\boldsymbol}

\makeatletter

\definecolor{mc}{HTML}{fa8caa}
\definecolor{dmc}{HTML}{d85a7a}
\definecolor{amc}{HTML}{6ec8f0}
\definecolor{damc}{HTML}{5790c9}
\hypersetup{colorlinks=true, allcolors=damc}

\author{%
  Qiyang Li \quad\quad Zhiyuan Zhou \quad\quad Charles Xu \quad\quad Sergey Levine \\
  UC Berkeley \\
  \href{https://data-inoculation.github.io}{\textcolor{dmc}{data-inoculation.github.io}}
}

\title{Robotic Data Inoculation}

\begin{document}

\begin{abstract}
Modern robot learning systems feed on high-quality demonstration data, and their performance hinges on coverage. Without good coverage, small errors compound and the policy can quickly go into out-of-distribution states from which it cannot recover. 
Yet there is no principled way of collecting high-coverage data, and data collection often relies on heuristic instructions to human operators. Purely online reinforcement learning (RL) explores at the cost of safety and sample efficiency, while DAgger-style interventions require the policy to fail by chance before a human can correct it.
In this paper, we ask instead whether we can modify how the robot behaves \textit{during data collection} to collect more useful data.
We propose \textcolor{dmc}{\textbf{Data Inoculation (DI)}}, a shared-control framework in which a human provide closed-loop teleoperation commands while a residual policy perturbs these commands. The residual policy is trained online, on the growing dataset, with the same reward-maximizing objective later used for policy extraction, so its perturbations steer data collection toward the deviations a policy trained on the data would make. 
These perturbations expose the robot to a wider range of states, while the human’s continuous corrections provide recovery demonstrations, ``inoculating'' the dataset against the learner's future mistakes while the human remains in control. On three challenging real-world robotic tasks, DI consistently improves or matches the success rate of policies trained with both behavioral cloning and offline RL, showing DI as an effective data collection strategy for improving downstream policy performance.
\end{abstract}

\maketitle

\begin{figure}[H]
    \centering
    \includegraphics[width=\linewidth]{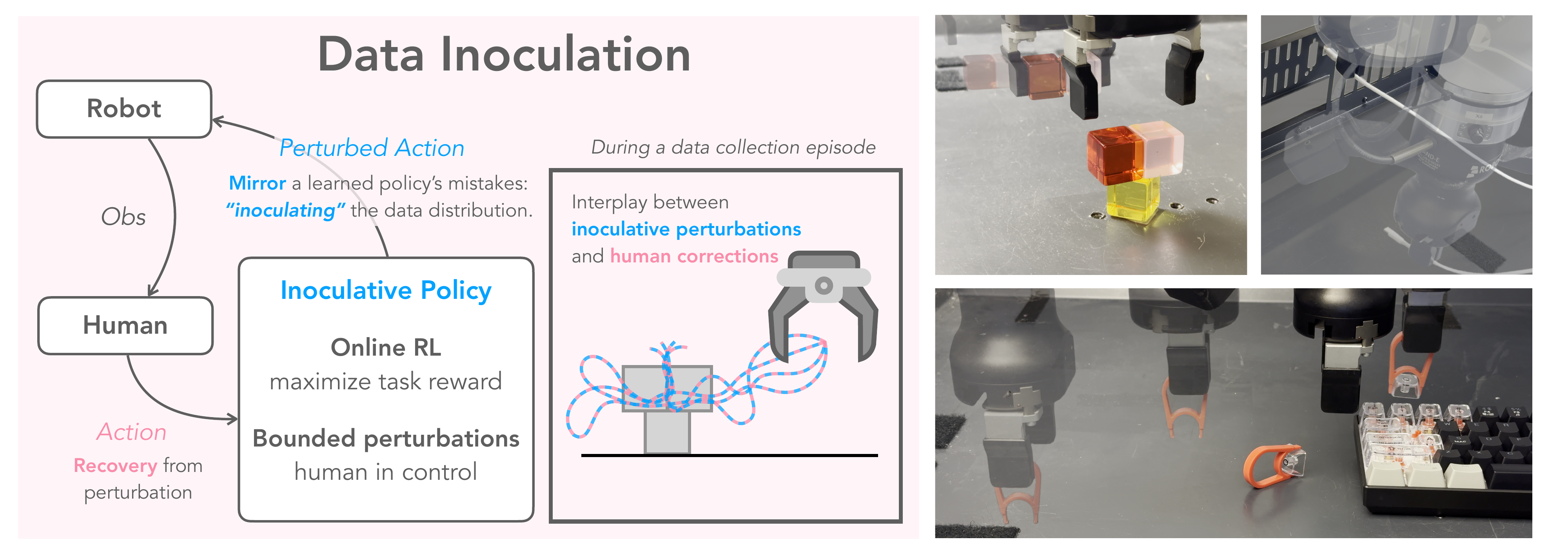}
    \caption{\textcolor{dmc}{\textbf{Data inoculation (DI). }} 
    \textit{Left}: During data collection, we train a residual inoculative policy with online RL to perturb the human's closed-loop teleoperation commands. The perturbations steer data collection towards the mistakes the policy trained on the data would make, and elicit human's corrective feedback, ``inoculating'' the dataset against such mistakes.
    \textit{Right}: We test DI on three challenging real-world robotic tasks that require precise manipulation, and DI consistently improve success rate of policies trained with behavioral cloning and offline RL.
    }
    \label{fig:main}
\end{figure}

\section{Introduction}
\label{sec:intro}

Modern robot learning systems have shown promise in acquiring complex manipulation skills (\emph{e.g.}, shirt-folding~\citep{pi05}, tea/juice-making~\citep{lwd}) from demonstration data. 
While there has been extensive recent research on the design of these models, the quality of the data the models are trained on is every bit as critical. Good data should have high coverage: it should contain a wide range of states to maximize the chance that the policy generalizes to the states that it actually visits at test time. In practice, coverage is determined by the scenes where data is collected and the behaviors of the teleoperators, which must strike a balance between consistency and variety. Improving data quality and coverage through task design and operator instruction is largely heuristic and ad hoc. Can we instead directly modify how the robot behaves \emph{during} data collection to provide for more useful data?

Existing solutions largely sit at the two extremes of how the robot is being controlled. Aside from the human-driven teleoperation which exhibits obvious drawbacks above, a learned policy may control the robot autonomously and collect data. While this allows us to run RL online which improves the policy upon its own experience overtime, real-world exploration can be sample inefficient and unsafe. DAgger-style methods partially address this by using learned policy to drive the robot until it makes a mistake, where a human intervenes and corrects it, often through teleoperation. This is reactive: each intervention requires a failure to occur naturally, and rare failure modes may take many rollouts to appear, and for many tasks (\emph{e.g.}, that require working with delicate objects that may break) failures cannot be corrected once they occur. 

In this paper, we argue that autonomous policy rollout and human control should \textit{not} be mutually exclusive, and a shared-control framework in which a learned agent and the human drive the robot together allows for better data collection. Our framework is simple: the human retains closed-loop teleoperation of the robot that supplies the demonstration signal as usual in the form of control commands, but these commands are perturbed by a residual policy, which is trained online on the growing dataset, in a way that mimics the mistakes a policy trained on the current data might make. We call this residual policy an \emph{inoculative policy} because it exposes data collection to controlled, human-corrected ``doses'' of the learner's expected mistakes, so that the policy learned from the collected data is already robust to them. The human can therefore correct policy-like deviations continuously throughout every data collection episode rather than only after a visible failure while ensuring safety because they are always in control of the robot. The resulting data collection strategy, which we call {\color{dmc}\textbf{Data Inoculation (DI)}} (\cref{fig:main}), produce data that are favourable for downstream policy extraction---it contains near-failure states paired with actions that allow recovery from these states, which is exactly the good coverage data that would help behavior cloning (BC) to learn the recovery behavior directly but also help offline RL to learn better value functions. 

Our main contribution is the new data collection strategy (DI). We evaluate DI on three challenging real-world robotic tasks and show that DI consistently improves or matches the success rate of the policies extracted from both behavior cloning (BC) and offline RL. These results demonstrate the benefits of proactively \emph{inoculating} data collection so that the policies become more robust at deployment time.

\section{Related Work}

\paragraph{Human-in-the-loop learning} methods address a common failure mode of imitation learning on passive data, where the learned policy may visit states different from the training distribution and suffers from compounding errors. They typically work by alternating between executing the current policy and collecting human correction data in the form of action re-labeling~\citep{ross2011dagger} or intervention trajectories~\citep{kelly2019hgdagger}. Interventions can be triggered automatically by the model~\citep{zhang2016query, hoque2021thriftydagger, menda2019ensembledagger} or initiated by human~\citep{kelly2019hgdagger,mandlekar2020human,luo2025precise,liu2025robot,hu2026rac,pi06}. In real-world robotics, intervention-based approaches are often more convenient and natural to human because the corrective trajectories are collected under human's continuous, closed-loop control of the robot. However, this also comes with a limitation: once the control is switched to human, the trajectory may quickly go out of the learned policy's state distribution. As a result, only the first few transitions right after the switching point provide direct supervision to the mistakes made by the policy. 
In contrast, rather than using a binary switching from the learned policy's distribution to the expert's distribution after intervention, our method constructs a controlled distribution by combining human demonstrator actions with residual actions from an \emph{inoculation policy}, allowing the human to provide corrections of the policy's mistakes throughout a data collection trajectory.
Prior work trains assistive residual policies in simulation, using a model trained on a fixed set of demonstrations to approximate human control, then deploys them with human operators~\citep{sha2026efficient, schaff2020residual}. 
In contrast, our method doesn't depend on simulation and directly train the residual inoculation policy online as data accumulates. This also yields a broader range of corrective demonstrations as updates to the residual policy expose humans to different errors during its training.

\paragraph{Data perturbation} approaches broaden the distribution of observations~\citep{tobin2017domain, laskin2020reinforcement, young2021visual, zhou2023nerf}, actions~\citep{zhou2023nerf, ke2024ccil, hoque2024intervengen, mandlekar2023mimicgen}, or even environment and its dynamics~\citep{huang2025adversarial, wang2024cyberdemo, peng2018sim, tan2018sim} during training so that the model is robust. These approaches typically directly perturb either a simulator (\emph{e.g.}, domain randomization) or previously collected data (\emph{e.g.}, data augmentation), but do not directly interact with how a human demonstrator collects trajectories, which our method does. The closest line of work injects zero-mean Gaussian noises during data collection, exposing human demonstrator to random mistakes and thereby eliciting recovery behaviors~\citep{laskey2017dart,oh2023bayesian,tahara2023disturbance}. In contrast, our method injects directed residual actions produced by a state-dependent \emph{inoculation policy} trained online to approximate the mistakes that a policy learned from the growing dataset would likely make. This allows the collected trajectories to target policy-specific failure modes like DAgger-style approaches, rather than relying on symmetric random perturbations. Furthermore, while prior approaches mostly focus on improving imitation learning performance, we additionally demonstrate that our technique allows for improved offline RL performance and the quality of learned value functions.

\section{Preliminaries}
\label{sec:background}
\textbf{Learning robot policy via reinforcement learning.} We consider a Markov decision process (MDP), $\mathcal{M} = (\mathcal{S}, \mathcal{A}, r, \gamma, \rho)$, with state space $\mathcal{S}$, action space $\mathcal{A}$, transition dynamics $T: \mathcal{S} \times \mathcal{A} \to \Delta_{\mathcal{S}}$, reward $r: \mathcal{S} \times \mathcal{A} \to \mathbb{R}$, discount factor $\gamma \in [0, 1)$, and an initial state distribution $\rho \in \Delta_{\mathcal{S}}$. The robot policy receives a partial observation of the state, defined by an observation model, $\Phi: \mathcal{S} \to \mathcal{O}$ (\emph{e.g.}, image observations, proprioceptive states, and the previous actions that are being executed on a robot). The performance of a policy $\pi : \mathcal{O} \to \Delta_{\mathcal{A}}$ is estimated by its $Q$-function, $Q^\pi(\mathbf s, \mathbf a) := \mathbb{E}_\pi\!\left[\sum_t \gamma^t r_t\right]$, averaged over the initial state distribution $\mathbf \sim \rho$. We use a standard goal-reaching sparse-reward formulation where $r = -1$ until the goal is reached. Optimizing $\pi$ against $\mathbb{E}_{\mathbf s \sim \rho}[Q^\pi(\mathbf s, \mathbf a)]$ corresponds to completing the task as fast as possible. Given a dataset $D$ of transitions $(\mathbf o, \mathbf a, \mathbf o', r)$, we may use the standard temporal-difference (TD) learning procedure where we approximate $Q^\pi(\mathbf s, \mathbf a)$ by $Q_\phi(\mathbf o = \Phi(\mathbf s), \mathbf a)$ and train it via the following TD loss:
\begin{align}
    L(\phi) = \mathbb{E}_{\mathbf a' \sim \pi(\cdot \mid \mathbf o')}\left[\left(Q_\phi(\mathbf o , \mathbf a) - r -\gamma Q_{\bar \phi}(\mathbf o', \mathbf a')\right)^2\right].
\end{align}
Meanwhile, we can train the policy to hill-climb on $\mathbb{E}_{\mathbf a \sim \pi(\cdot \mid \mathbf o)}\left[Q_\phi(\mathbf o, \mathbf a)\right]$ to improve its goal-reaching capability.

\textbf{Vanilla human data collection.} A typical human data collection episode goes like this: a human teleoperator drives the robot in closed loop, giving a control command $\mathbf{a}_t$ at every step for the robot to execute, while continuously observing the scene to provide any immediate command corrections needed for task completion. The episode terminates on task success, non-recoverable failure (\emph{e.g.}, objects fallen off the table), or a time limit. Throughout the episode, we record a trajectory of control commands and the observations (\emph{e.g.}, by reading off the proprioceptive states provided by the robot and the images from camera(s) attached to the robot) that are being synchronized to the same time steps (\emph{i.e.}, $\{\mathbf{a}_t, \mathbf{o}_t\}_{t=1}^T$). Since a proficient human teleoperator is often fast and consistent, the states in such data are often narrow and lack recovery. The data may teach the policy how to \emph{proceed} from states an expert visits, but lacks supervision at the states an imperfect learned policy might actually visit at deployment when deviating from the expert trajectories, nor corrective commands that may recover from these deviations.

\section{Data Inoculation}
\label{sec:inoculation}

In this section, we describe in detail how we use a residual policy in a shared-control framework to continuously inject controlled perturbations such that the collected data is favorable for downstream policy training. 
We start by discussing how a residual policy alongside closed-loop human control can address the failure mode of vanilla human teleop data collection, and then detail how such a residual policy can be learned online during data collection with online RL.

\textbf{Method overview.} 
As discussed in \Cref{sec:background}, vanilla human data collection can suffer from lack of corrective and recovery demonstrations.
DI solves this by intelligently modifying what the robot executes. At each step, an \emph{inoculative policy} $\pi_\theta$ takes the current observation $\mathbf o_t$ and the teleoperator's live command $\mathbf a_t$, and generates a unit $L_\infty$ bounded residual perturbation $\Delta_t \sim \pi_\theta(\cdot \mid \mathbf{o}_t, \mathbf{a}_t)$ so that the executed command $\tilde{\mathbf{a}}_t \leftarrow \mathbf{a}_t + \alpha \Delta_t$ deviates from the teleoperator's by at most $\alpha$ in $L_\infty$-norm. The teleoperator can then immediately react to the deviation and counter-corrects it through subsequent commands. Every perturbation is then naturally paired with human's follow-up recovery in the trajectory. 

Injecting purely random noises without any guidance inefficient, because perturbations along axes that barely influence task success (\emph{e.g.}, yaw rotation in a simple cube pick-and-place task) are uninformative to the downstream learner. Intuitively, the data collection budget is best spent on the critical perturbation axes that influence the task success the most, and elicit recovery behavior as a consequence of these perturbations. However, finding informative perturbations is not simple as they are highly task-dependent and non-stationary, requiring a form of fast adaptation that vanilla zero-mean random noises (as often used in prior work~\citep{laskey2017dart}) cannot provide. How should we best implement an algorithm that can quickly adapt and find the perturbation axes that are most informative? It turns out that we can achieve this altogether by simply optimizing the inoculative policy along with human data collection using standard online RL to \emph{maximize task success}. 

In particular, we train our inoculative policy using standard TD-based off-policy RL, concurrently with collection, on the growing dataset from the data collection itself. While the data collection loop runs on the robot at the normal control frequency, an asynchronous RL training loop updates $\pi_\theta$ on the incoming data stream to maximize task success rate---or in other words, to maximize its ability to \emph{assist} the teleoperator in completing task. This may seem counter-intuitive: how can a \emph{cooperative} policy produce informative perturbations and elicit recovery behaviors? 

This answer is two-fold. First of all, when the inoculative policy is optimized using the same reward-maximizing objective as a downstream offline RL algorithm later uses for policy extraction, its deviations are expected to closely mirror those a downstream learned policy would produce. Furthermore, early in collection, the critic is barely trained and inaccurate which causes the residual policy to exploit the critic's errors, steering the robot into suboptimal states that require explicit recovery behaviors. Aside from eliciting recovery behaviors, our inoculative policy can sometimes also improve data collection efficiency (as we will show in \cref{fig:t-cube-success}) because it is optimized to cooperate with human to improve task success rates. 

We have thus far described how training our residual inoculative policy with online RL leads to more informative data collection with the teleoperator maintaining closed-loop control of the robot. Next, we will describe important implementation details that allow efficient learning of our inoculative policy.

\textbf{Learning the inoculative residual policy online.} 
We train a critic $Q_\phi(\mathbf o, \tilde{\mathbf a})$ that estimates the value of the perturbed action directly rather than the value of the residual action $\Delta$. This allows us to update $Q_\phi$ on any off-policy data regardless whether the action has been perturbed (in practice we collection some portion of our data un-perturbed). To establish the training loss for $Q_\phi$, we start from the standard temporal-difference (TD) loss constructed from a transition $(\mathbf o, \mathbf a, \mathbf o')$, where $\tilde{\mathbf{a}}'$ is the command perturbed by the current $\pi_\theta$ on top of the human teleoperation command that would have been given at observation $\mathbf o'$:
\begin{align}
    L(\phi) = \big(Q_\phi(\mathbf o, \tilde{\mathbf{a}}) - r - \gamma Q_{\bar \phi}(\mathbf o', \tilde{\mathbf{a}}')\big)^2.
\end{align}
However, this requires $\tilde{\mathbf a}'$ which cannot be obtained from the residual policy $\pi_\theta$ alone because it needs a clean action $\mathbf{a}'$ as input that the transition does not have. To approximate $\mathbf{a}'$, we additionally train a behavior-cloning (BC) flow model $\pi_\beta$ that imitates the human teleoperation commands, trained on the same growing dataset from the data collection itself.~\footnote{We could use $\mathbf{a}'$ recorded in the trajectory instead of the flow BC approximation, but this creates an unnecessary non-Markovian dependency on $\mathbf o, \mathbf a$ which may hurt learning in theory, so we did not pursue in our experiments.} With such approximation, our modified TD loss is therefore
\begin{align}
    L(\phi) = \big(Q_\phi(\mathbf o, \tilde{\mathbf{a}}) - r - \gamma Q_{\bar \phi}(\mathbf o', \mathbf{a}' +
      \alpha\Delta')\big)^2, \quad\quad \mathbf a' \sim \pi_\beta(\cdot\mid \mathbf o') \quad \Delta'\sim \pi_\theta(\cdot \mid \mathbf o', \mathbf a').
\end{align}
We can then simultaneously use $Q_\phi$ to train our inoculative policy $\pi_\theta$ via the following policy loss:
\begin{align}
    L(\theta) = -\mathbb{E}_{\Delta \sim \pi_\theta(\cdot \mid \mathbf o,
      \mathbf a)}\big[Q_\phi(\mathbf{o}, \mathbf{a} +
      \alpha\Delta) - \lambda \lVert \Delta \rVert_2^2\big],
\end{align}
where $\lambda$ is a tunable parameter that additionally penalizes the magnitude of the perturbation which we find to be helpful for environments that require very precise commands (\emph{e.g.}, \texttt{t-cube}).  

\textbf{Delay-aware action chunking policy learning.} Since the inoculative policy takes the human command as input and adds perturbation on top of it, the executed command needs to be delayed at least the latency of policy inference. To make this work in practice, we adopt a delay-aware learning scheme where we always execute commands a \emph{fixed} interval $\delta$ after the observation and human command. We keep a constant $\delta$ and evaluate the learned policy with the same fixed delay. To make sure $\mathbf o_t$ contains enough information about $\mathbf s_{t+\delta}$ ($t+\delta$ is when the action $\mathbf a_t$ is actually executed), we include both the observation at $t$, and the action being executed from $t$ to $t +\delta$:
\begin{align}
    \mathbf o_t = ( I_t, p_t, \mathbf a_{t-\Delta t} ),
\end{align}
where $I_t$ is the encoding (using frozen pre-trained ResNet-18 following \citet{luo2024serl}) of image observations at $t$, $p_t$ is the proprioceptive state at $t$, and $\mathbf a_{t-\Delta t}$ being the action executed at the previous time step.

\textbf{Other practical considerations.} In our early experiments, we find that the changing perturbation direction too frequently can make the human hard to react to the perturbation effectively at all. In practice, we only predict the perturbation (the action residual) once every 0.5 seconds (5 control time steps with a frequency of 10 Hz). For all 5 time steps, we use the same perturbation added on top of $\mathbf a_t$ so that the teleoperator has a chance to correct any mistake that the perturbation induces. For our policy, we train an action chunking policy with a horizon of 5, where it predicts a sequence of 5 actions at a time and then execute at the control frequency with a fixed delay $\delta$. To extract this action chunking policy from dataset, we follow the BFN variant of Q-chunking~\citep{li2025qc} with a best-of-N of $N=4$. 

\begingroup
\setlength{\algomargin}{0pt}
\SetAlFnt{\small}
\begin{algorithm2e}[H]
\caption{Data Inoculation} \label{alg:cme}
\SetKwInput{Input}{Input}
\SetKwInput{Output}{Output}
\SetKwFor{RepTimes}{repeat}{times}{end}
\SetKwBlock{CollectorBlk}{\textcolor{damc}{Async Data Collection Loop:}}{}
\SetKwBlock{LearnerBlk}{\textcolor{dmc}{Async Training Loop:}}{}
\DontPrintSemicolon

\Input{Residual scale $\alpha$; residual penalty $\lambda$}
\Output{Inoculated demonstration dataset $D$}
\vspace{0.25em}

$D \leftarrow \emptyset$;\quad Initialize residual policy
$\pi_\theta$, critic $Q_\phi$, and BC flow policy $\pi_\beta$\;
\vspace{0.6em}
\setlength{\linewidth}{\hsize}
\noindent
\begin{minipage}[t]{0.41\linewidth}
\raggedright
\CollectorBlk{
\vspace{0.5em}
  $\mathbf{o}_0 \leftarrow$ initial observation after reset\;
\vspace{0.5em}
  \While{$t \in \{0, 1, \cdots\}$}{
  \vspace{0.5em}
    Observe $\mathbf{a}_t$ from human teleop\;
    \vspace{0.5em}
    $\Delta_t \sim \pi_\theta(\cdot \mid \mathbf{o}_t, \mathbf{a}_{t})$\;
    \vspace{0.5em}
    Execute $\tilde{\mathbf{a}}_t = \mathbf{a}_t + \alpha \Delta_t$\;
    \vspace{0.5em}
    Observe $\mathbf{o}_{t+1}$ and $r_t$\;
    \vspace{0.5em}
    $D \leftarrow D \cup
      \{(\mathbf{o}_t, \mathbf{a}_t,\tilde{\mathbf{a}}_t,\mathbf{o}_{t+1}, r_t)\}$\;
  }
}
\end{minipage}\hfill\begin{minipage}[t]{0.56\linewidth}
\raggedright
\LearnerBlk{
\vspace{0.5em}
  Sample a training batch $(\mathbf{o}, \mathbf{a}, \tilde{\mathbf{a}},\mathbf{o}', r) \sim D$\;
  \vspace{0.5em}
  Update $Q_\phi$ with the following TD loss:
  \[
    \adjustbox{max width=\hsize}{$\displaystyle
      L(\phi) = \big(Q_\phi(\mathbf o, \tilde{\mathbf{a}}) - r
      - \gamma Q_{\bar \phi}(\mathbf o', \mathbf{a}' + \alpha\Delta')\big)^2,
    $}
  \]
  where $\mathbf{a}' \sim \pi_\beta(\cdot \mid \mathbf o')$,
  $\Delta' \sim \pi_\theta(\cdot \mid \mathbf o', \mathbf a')$\;
  \vspace{0.5em}
  Update $\pi_\theta$ with the following policy loss:
  \[
    \adjustbox{max width=\hsize}{$\displaystyle
      L(\theta) = -\mathbb{E}_{\Delta \sim \pi_\theta(\cdot \mid \mathbf o, \mathbf a)}
      \big[Q_\phi(\mathbf{o}, \mathbf{a} + \alpha\Delta)
      - \lambda \lVert \Delta \rVert_2^2\big],
    $}
  \]
  Update $\pi_\beta$ with flow-matching loss using $(\mathbf o, \mathbf a)$\;
}
\end{minipage}\;
\end{algorithm2e}
\endgroup

\section{Experimental Setup}
\label{sec:experimental-setup}

We empirically evaluate our method across three different challenging precise manipulation tasks in the real world. We first describe the experimental setup, then the results in \Cref{sec:results}.

\paragraph{Robot setup and tasks.}
We evaluate our method on three real-robot tasks that require precise manipulation. We focus on these tasks because successful executions require precise control of the robot, making it a challenging testbed for data collection and policy learning. 
We use a Franka robot arm with a Robotiq gripper and two wrist cameras attached near the end effector. We use a spacemouse to control the robot's end effector position and rotation during data collection at a frequency of $10$Hz.
Illustrations of the three tasks are shown in \Cref{fig:main}, and we describe them in detail below:
\begin{enumerate}
    \item \texttt{t-cube} requires the robot to pick up two colored cubes and carefully balance them on top of another yellow cube, resulting in a T-shape~\citep{ghugare2025builderbench}. While this seems like a simple pick and place task, this in fact requires very precise manipulation skills and is arguably the most demanding task of the three: the robot needs to stack the two top cubes diagonally on top of the yellow cube with the centers of the three cubes roughly aligned, and any misalignment would cause one or two of the top cubes to fall off.
    \item \texttt{cable} is a precision insertion task where the robot needs to pick up a USB-C cable and insert it into a port hole on a desktop machine that is barely bigger than the cable itself. Without changing the difficulty of the task, we only use the side plate of the desktop machine for ease of task setup.
    \item \texttt{keycap} tasks the robot to use a keycap puller to remove the Q keycap from a keyboard. The robot must pick up the tool, locate the Q key, position the puller over the keycap, and lock its two hooks onto both sides of the keycap. Successful removal requires precise, directionally controlled force rather than a large brute force pull, which could cause the keycap to fly off. The robot must therefore learn a sequence of small lateral wiggling motions to loosen the keycap before pulling it out and placing it on the table.
\end{enumerate}

\paragraph{Data collection procedure.}
Since data collection through tele-operation is unavoidably a subjective process and can vary across different operators and collection trials, we conduct experiments on all tasks and data collection strategies with multiple operators and trials, which we refer to as ``seeds''. For all experiments, three different operators conduct data collection for a total of $9$ seeds for the \texttt{t-cube} task and $5$ seeds for the other two tasks. We collect more seeds for \texttt{t-cube} because it is the most difficult task.
Before data collection, each operator spent time to practice on the task to achieve a sufficient level of expertise before collecting any data.

\paragraph{Policy learning and evaluation.}
The observations for the policy consist of the two wrist RGB images, as well as proprioceptive state consisting of the end-effector position, rotation, and gripper position.
The policy outputs a $5D$ action as the command, consisting of the delta end effector position, the delta yaw, and the absolute gripper position.
For policy training, including both the residual policy used in DI and the downstream policy trained on the collected dataset, we use action chunking~\citep{li2025qc} with a chunk size of $h=5$. In this setting, the policy learns to output a sequence of $h$ atomic actions and execute them in open loop.
To evaluate policy performance, we train separate policies for each seed of data collection, and measure success rate over $20$ rollouts.

\paragraph{Comparisons.}
We compare our data collection framework DI against two baseline data collection strategies.
The first, \textbf{\textit{clean}}, refers to collecting data with pure teleoperation, where the control commands from the operator is passed ``cleanly'' without modification to the robot for closed-loop execution.
The second, \textbf{\textit{random}}, adds uniform random noise to the control command from the teleoperator before executing on the robot. The noise is sampled uniformly from $[-c, c]$ ($c=0.05$ for \texttt{t-cube/cable}, $c=0.1$ for \texttt{keycap}), where $0.1$ in the action space translates to $3$\,mm in end effector position and $0.01$\,rad in rotation. Similar to DI, we do not add any random noise to the gripper dimension of the action because random opening and closing of the gripper is usually harmful for task progress. In addition, we apply the same random noise for a sequence of $h=5$ actions, which is the same horizon we use for action chunk learning for the residual policy in DI.

We evaluate the quality of the collected data through downstream policy performance under both imitation learning and offline RL. For imitation learning, we train a flow-matching policy using behavior cloning (BC). For offline RL, we use BFN~\citep{li2025qc},
which uses the learned $Q$-function to perform rejection sampling with the BC policy: the BC policy proposes $N=4$ candidate actions, and the action with the highest predicted $Q$-value is selected.

\section{Results}
\label{sec:results}

In this section, we conduct real-robot experiments to study the effectiveness of DI for data collection and its impact on downstream policy learning. We study the following research questions:

\medskip
\noindent\textbf{Q1. Does DI improve policy performance for BC and offline RL?}

\begin{figure}[t]
    \centering
    \includegraphics[width=\linewidth]{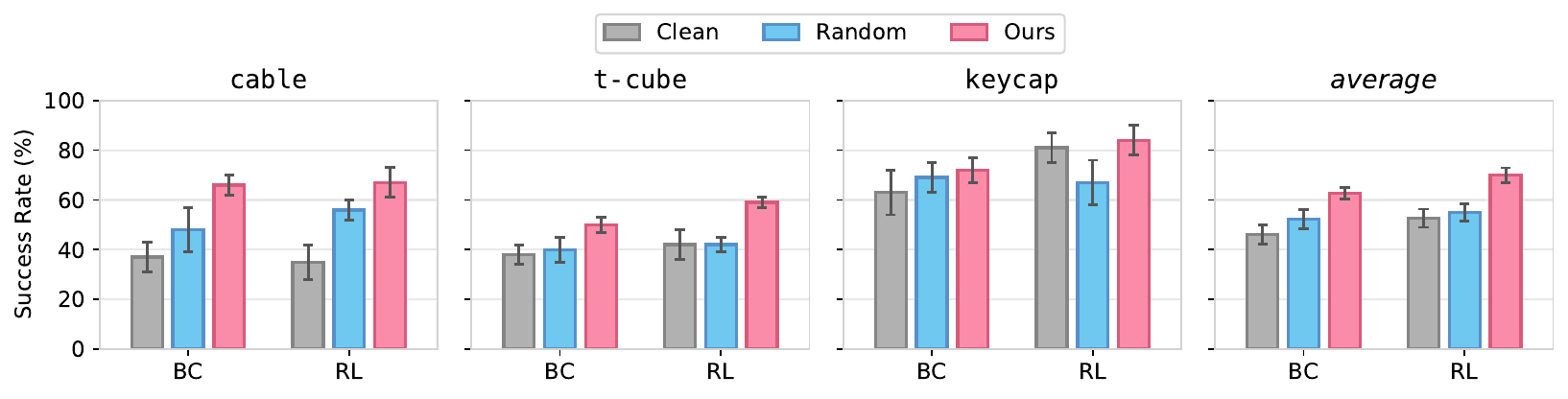}
    \caption{\textbf{Task success rates.} The results for \texttt{\{t-cube/cable/keycap\}} are averaged over \texttt{\{9/5/5\}} seeds with 20 evaluation trials each. Error bars indicate the standard errors. Full table in \cref{tab:success-rates,tab:cube-per-seed,tab:cable-per-seed,tab:keycap-per-seed}.}
    \label{fig:success-rates}
\end{figure}

\smallskip
\noindent First, we investigate whether data collection with DI can improve the performance of the resulting policies, as compared to \textit{clean} and \textit{random} data collection.
Since different policy collection methods have different degrees of optimality, we evaluate policies trained with both behavioral cloning and offline RL, as described in \Cref{sec:experimental-setup}.
\Cref{fig:success-rates} shows the success rates of both BC and RL policies trained with each data collection method on each of the three real-robot tasks. Overall, our DI data collection method leads to the highest success rate policies, regardless of whether we train with BC or offline RL. This shows that the residual inoculative policy is able to help collect more useful data for downstream learning.
Compared to \textit{clean} human demonstrations, data collection with DI significantly improves policy success rate on \texttt{cable} and \texttt{t-cube}, and is similar or slightly better on \texttt{keycap}.
The \texttt{keycap} task shows similar performance across all three data collection methods, perhaps because the task is less demanding in terms of precise manipulation skills and recovery behavior (supported by high success rate of clean human demonstration in \Cref{tab:collection-success}),
and so plain human demonstration is mostly sufficient.
Compared to data collection with injected random noise, which only leads to policy performance improvement on \texttt{cable} and can hurt performance on \texttt{keycap}, DI consistently improves policy performance for both RL and BC policies. This demonstrates that the residual policy learns meaningful state-dependent noise to inject into data collection.

\medskip
\noindent\textbf{Q2. How does DI affect human demonstrator's success rate?}

\begin{figure}[h]
    \centering
        \includegraphics[width=\linewidth]{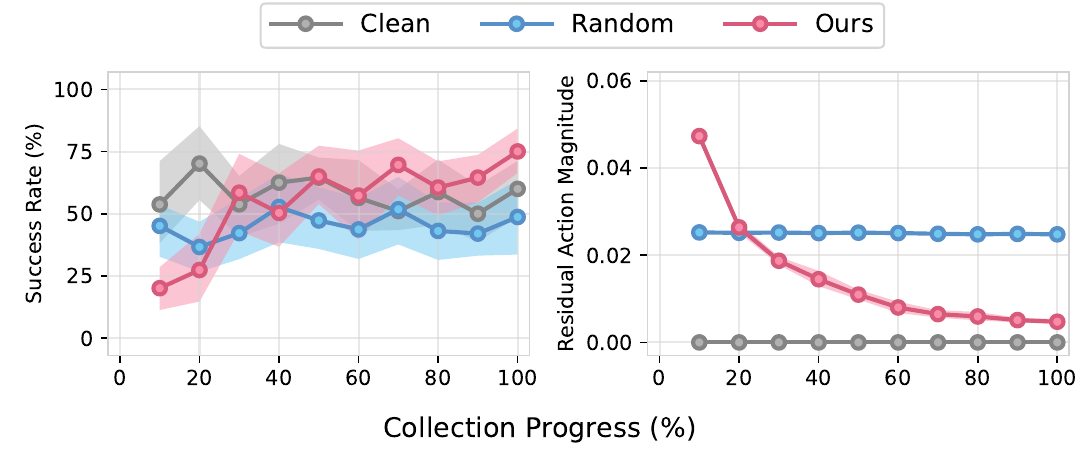}
    \caption{\textbf{Success rate over the course of data collection.} DI starts at a lower success rate when the residual policy is untrained, but eventually reaches a higher success rate than clean and random collection. 
    }
    \label{fig:t-cube-success}
\end{figure}

\smallskip
\noindent Given that DI artificially injects a perturbation into the human demonstration, it is natural to ask how DI affects the human demonstrator's success rate during data collection. In \Cref{fig:t-cube-success}, we measure the success rate over the course of data collection for the three methods. For both \textit{clean} and \textit{random} data collection, the success rate remains mostly constant.
For DI, the success rate starts lower than its counterparts. This is expected since the residual inoculative policy outputs near random actions at the beginning of data collection, and this could negatively impact the success rate initially.
However, the success rate of DI quickly rises through data collection, matching the success rates of \textit{clean} and \textit{random} roughly a quarter way into the whole collection session and eventually achieves high success rate than its counterparts. This demonstrates that the residual inoculative policy can quickly learn useful behavior, and can \textit{assist} the human demonstrator to collect higher success rate data for precise manipulation tasks.
In contrast, collecting with a random noise almost always decrease success rate over the course of data collection.
We also measure the magnitude of the residual action in \Cref{fig:t-cube-success}, and found that the magitude of DI residual action consistently decreases over the course of collection, showing that the residual policy shift from outputting random noise to only useful small deviations from human demonstration.

\medskip
\noindent\textbf{Q3. Does DI Induce Useful Corrective Demonstrations from Human Operator?}

\begin{figure}[h]
    \centering
    \includegraphics[width=0.8\linewidth]{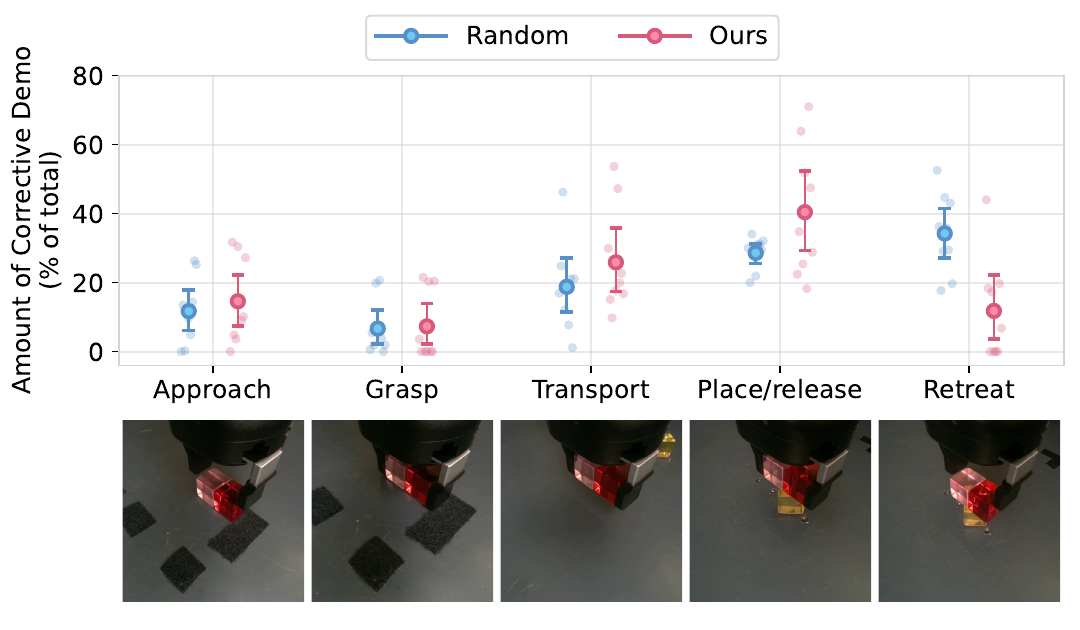}
    \caption{\textbf{Amount of human corrective demonstrations over different phases}. DI elicits the most corrections at the most critical phase of the \texttt{t-cube} task.}
    \label{fig:recovery-analysis}
\end{figure}

\smallskip
\noindent Since we observe above that DI improves downstream policy performance, we next seek to understand more mechanistically how DI leads to such improvement. Intuitively, the residual inoculation policy could perturb the robot into an unfamiliar state and elicit corrective demonstration from the human operator.
To test this intuition, we compare where the corrective demonstrations occur along trajectories collected with DI versus randomly injected noise.
We focus our analysis on the \texttt{t-cube} task in the following because it is the most precise task, and therefore the most revealing on corrective behavior. We divide the task into $5$ phases using rough heuristics: approach (start to close the gripper), grasp (from closing the gripper until $\geq 1$cm lift), transport (lift to final descent), place/release (final descent through place and release), and retreat (after release). Visualizations fo the five phases are in \Cref{fig:recovery-analysis}.
Then, we quantify the amount of corrective demonstration per phase by weighting the DI residual intervention in this phase by the strength of corrective demonstration.
Specifically, if we denote the residual action as $\Delta_t$, the human teleoperation command $\mathbf{a}_t$, the chunked corrective direction is $\Delta\mathbf{a} = \mathbf{a}_{t+H} - \mathbf{a}_t$.
We define the \textit{strength} of corrective demonstration to be
$c_s =
\max\left(
0,\,
-\frac{\sum_{t} \Delta_t^\top \Delta\mathbf{a}_t}
{\sum_{t}\lVert \Delta_t\rVert_2^2}
\right).
$
Intuitively, the numerator measures whether the human command changes in the direction opposite to the residual over the next $H=5$ control steps. Opposing changes produce a positive gain; aligned changes produce a negative value that is clipped to zero.
The denominator serves as a normalizing constant.
We then use $c_s$ to weight the total residual magnitude in this phase, quantifying the \textit{amount} of corrective demonstration:
$c_a = c_s \sum_{t}||\Delta_t||_1.$
\Cref{fig:recovery-analysis} shows amount of corrective demonstration ($c_a$) in each phase as a percentage of the total amount.
The result shows that most of the corrective behavior during DI collection happens during the place/release phase, the most difficult and central phase of \texttt{t-cube}. Failures in \texttt{t-cube} happen almost exclusively in this phase because any imprecise placement or gripper release can knock over the cubes. As such, this is the most important phase from which to collect corrective behavior.
In contrast, many corrective demonstration for random data collection happen during the retreat phase, which is relatively straight forward and does not need much corrections.
This shows that by learning a residual policy online that adapts to the task, DI is able to collect more useful corrective demonstrations.

\section{Discussion}
\label{sec:discuss}

In this paper we introduce Data Inoculation, a data collection strategy that trains a residual RL policy online to aid human teleoperated data collection in a shared-control framework. The residual inoculative policy perturbs the human's control of the robot to a wider range of states and can elicit recovery demonstrations in critical phases of the task, and eventually lead to higher data collection success rate as the residual policy is trained. Data collected with DI also improves the success rate of downstream policy learning.
Our work serves as a step towards a new paradigm of robot learning where data collection is shared between humans and autonomous policies, each helping the other. 
Several challenges remain and is promising for future work. First, our approach trains the residual policy from scratch using online RL, which offers near-random perturbations at the beginning of training. A natural next step is to develop mechanisms for directed exploration that can more quickly provide useful perturbations, such as boostrapping training from an existing offline dataset.
Second, our approach assume the human operator to be fully in-the-loop during data collection, which is expensive in human time. One interesting direction is to explore whether we can automate parts of data collection to be autonomous gradually during training and leave only important correction for human operators.

\section{Acknowledgments}
This research was supported in part by ONR under N00014-25-1-2060. We would like to thank Seohong Park for the name of the method.

\bibliographystyle{iclr2027_conference}
\bibliography{main}

\newpage
\appendix
\section{Compute Resource and Robot Hardware}
\label{appendix:resources}
We run all our experiments on a local workstation with two GPUs (RTX-4090). The first GPU trains the models ($\pi_\beta, \pi_\theta, Q_\phi$) and periodically save the model weights to disk, and the second GPU loads the model weights periodically. We use Franka Emika Panda robot arm with the impedance controller from the SERL~\citep{luo2024serl}. To teleoperate the robot, we use 3Dconnexion spacemouse. 

\section{Task Details}
\label{appendix:tasks}

\begin{figure}[H]
    \centering

    \begin{minipage}[c]{0.158\linewidth}
        \centering
        \phantom{\textbf{Task}}
    \end{minipage}
    \hfill
    \begin{minipage}[c]{0.52\linewidth}
        \centering
        \large\bfseries Task Trajectory
    \end{minipage}
    \hfill
    \begin{minipage}[c]{0.293\linewidth}
        \centering
        \large\bfseries Success State
    \end{minipage}

    \vspace{1mm}

    \begin{minipage}[c]{0.158\linewidth}
        \centering
        \Large\bfseries \texttt{t-cube}
    \end{minipage}
    \hfill
    \begin{minipage}[c]{0.52\linewidth}
        \includegraphics[width=\linewidth]{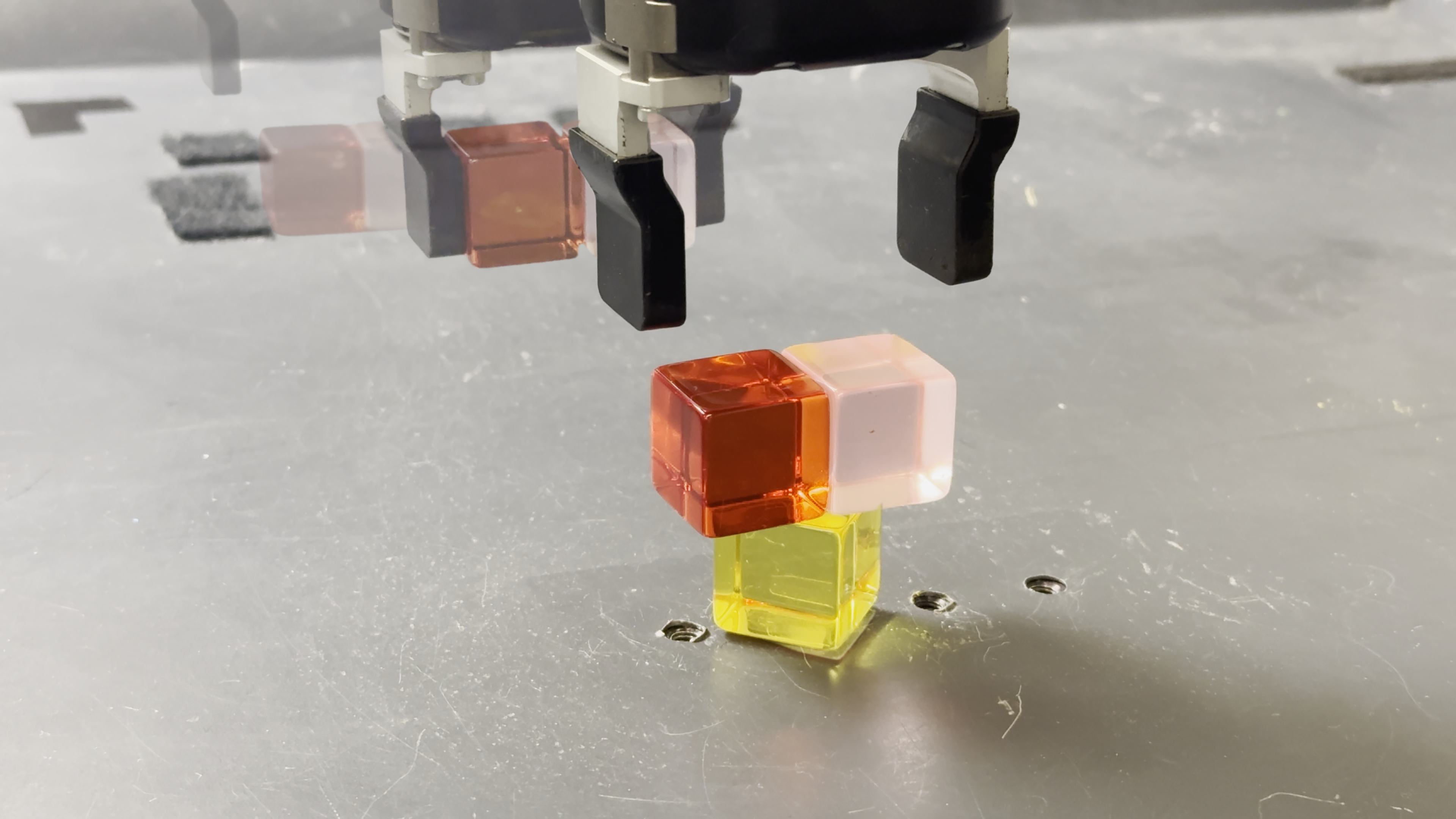}
    \end{minipage}
    \hfill
    \begin{minipage}[c]{0.293\linewidth}
        \includegraphics[width=\linewidth]{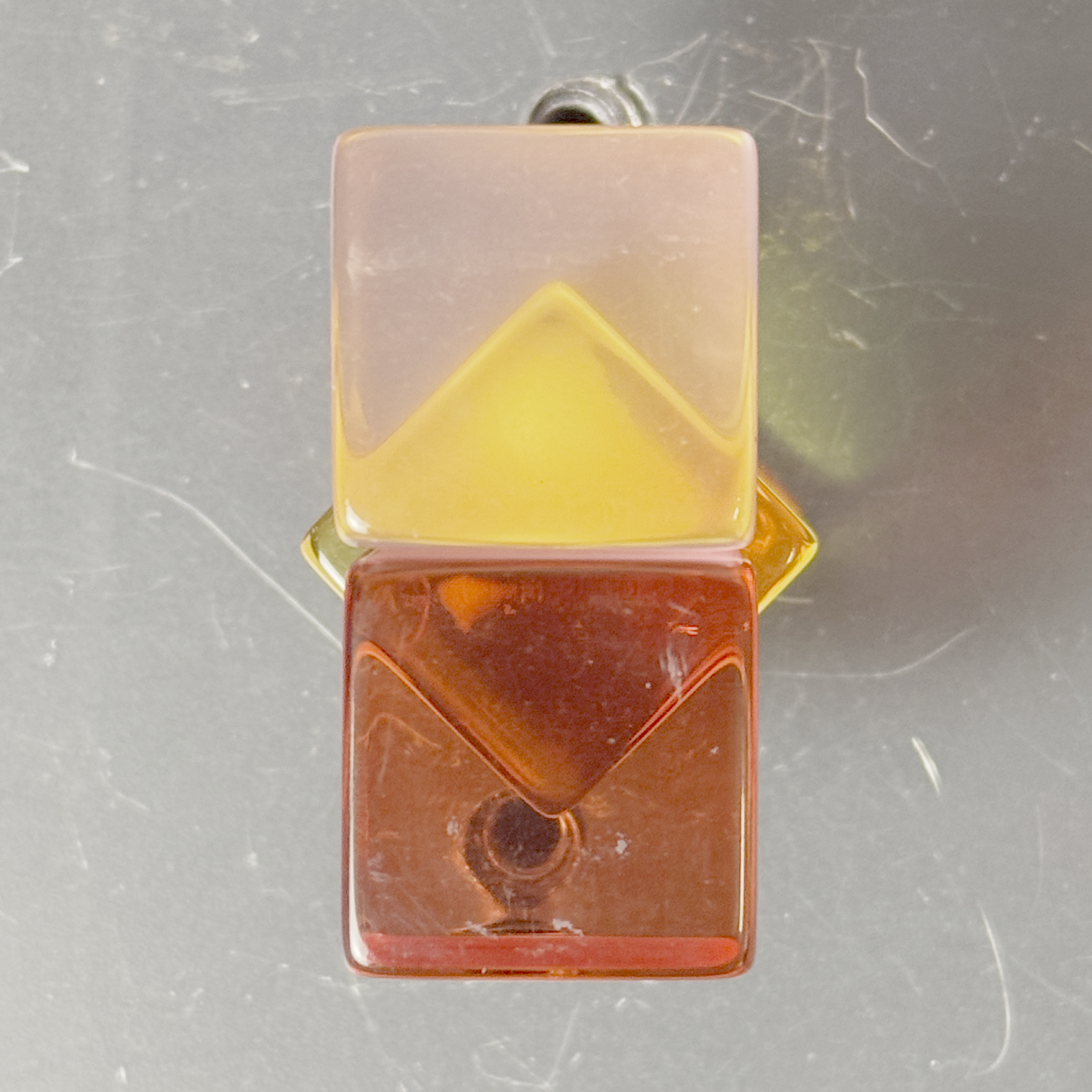}
    \end{minipage}

    \vspace{1mm}

    \begin{minipage}[c]{0.158\linewidth}
        \centering
        \Large\bfseries \texttt{cable}
    \end{minipage}
    \hfill
    \begin{minipage}[c]{0.52\linewidth}
        \includegraphics[width=\linewidth]{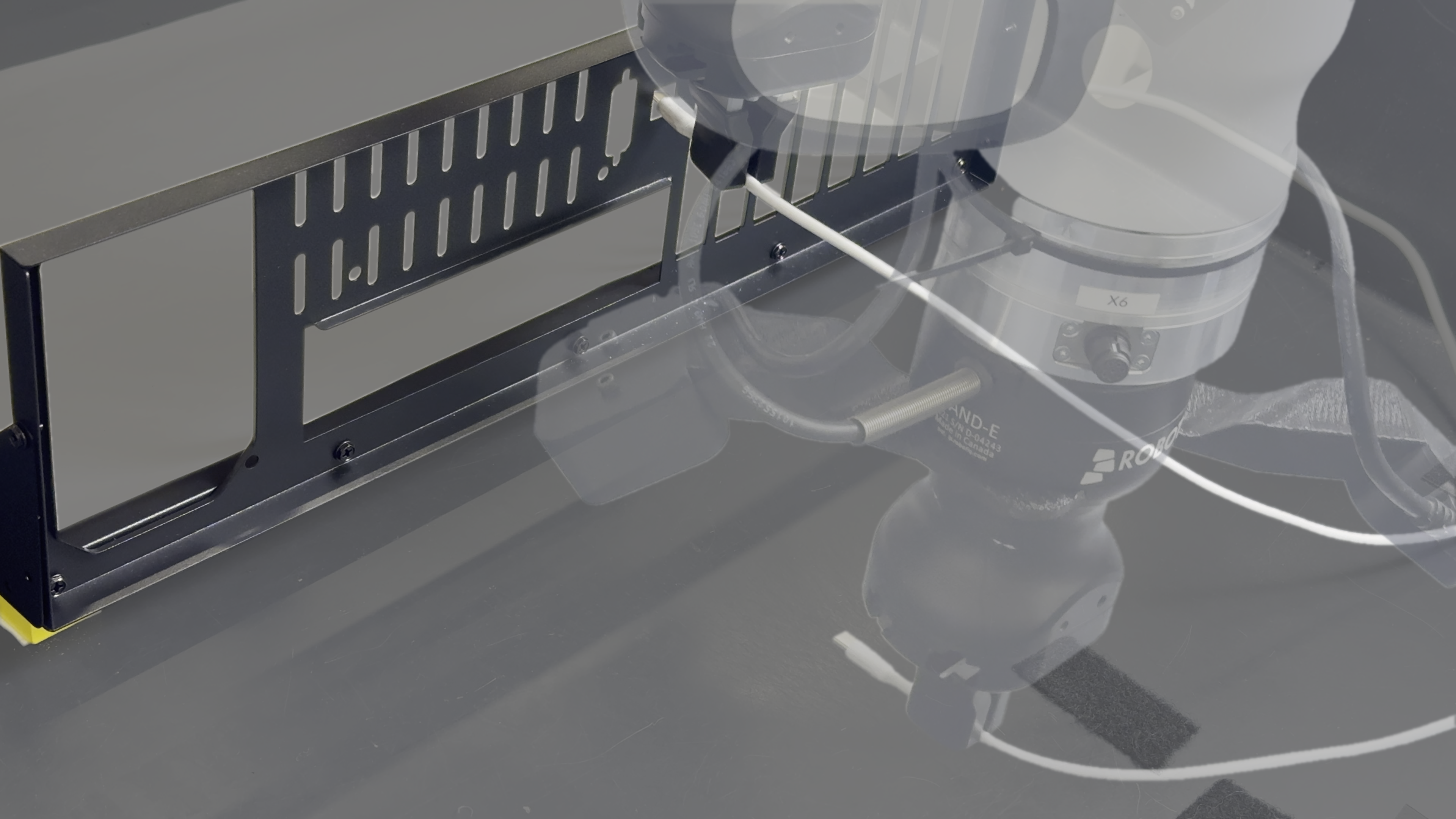}
    \end{minipage}
    \hfill
    \begin{minipage}[c]{0.293\linewidth}
        \includegraphics[width=\linewidth]{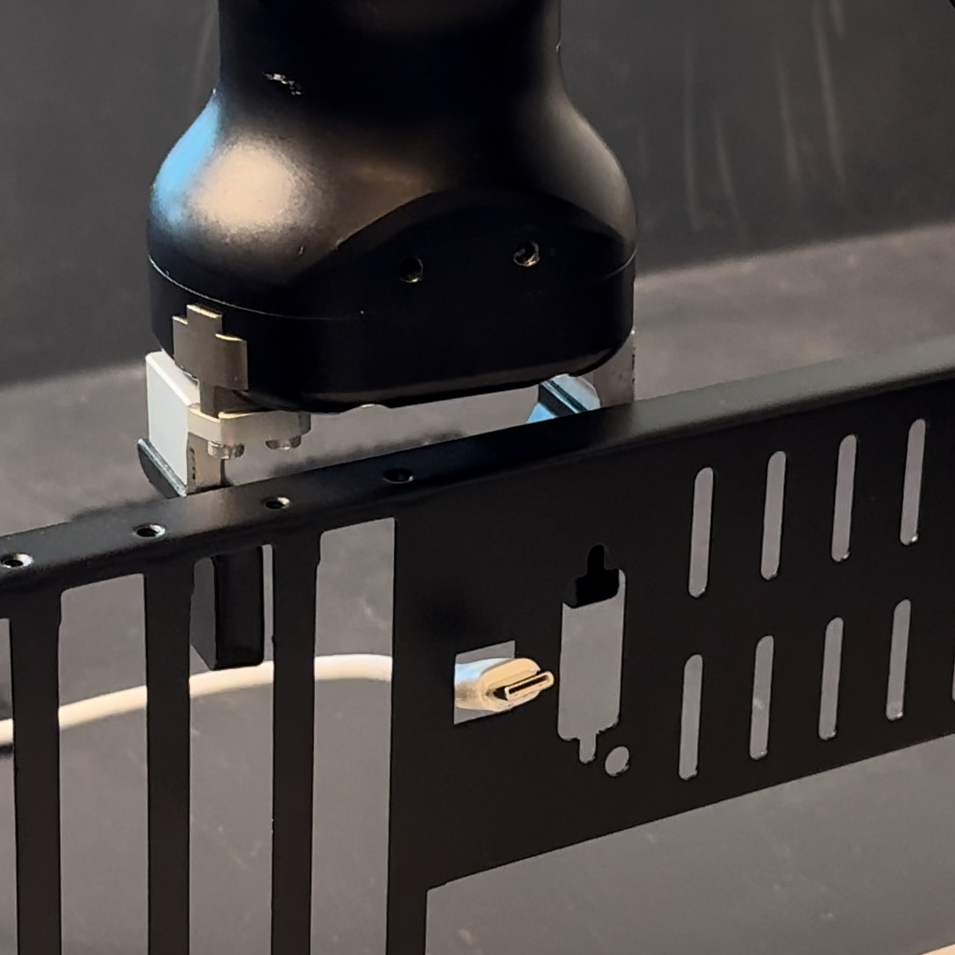}
    \end{minipage}

    \vspace{1mm}

    \begin{minipage}[c]{0.158\linewidth}
        \centering
        \Large\bfseries \texttt{keycap}
    \end{minipage}
    \hfill
    \begin{minipage}[c]{0.52\linewidth}
        \includegraphics[width=\linewidth]{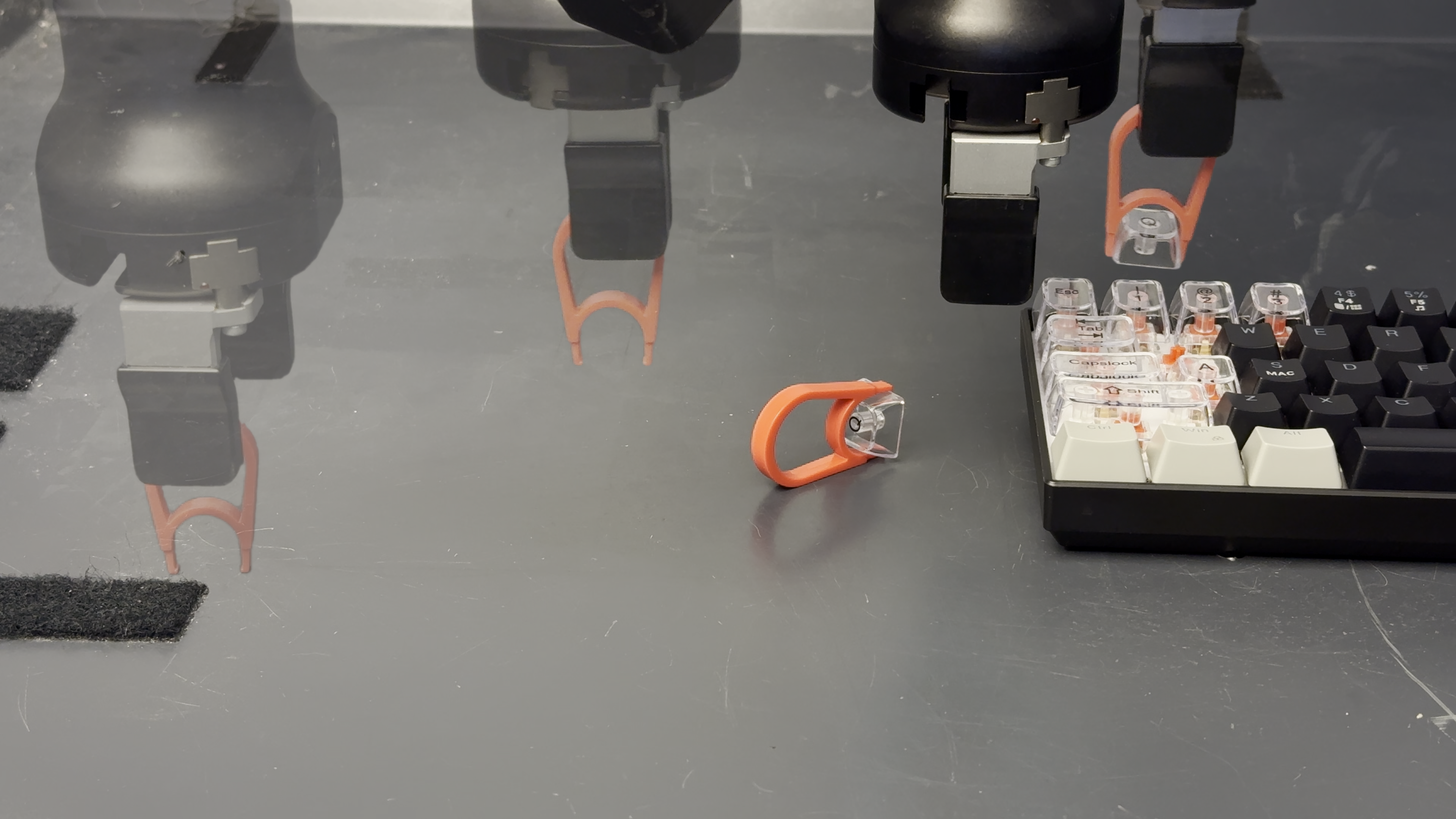}
    \end{minipage}
    \hfill
    \begin{minipage}[c]{0.293\linewidth}
        \includegraphics[width=\linewidth]{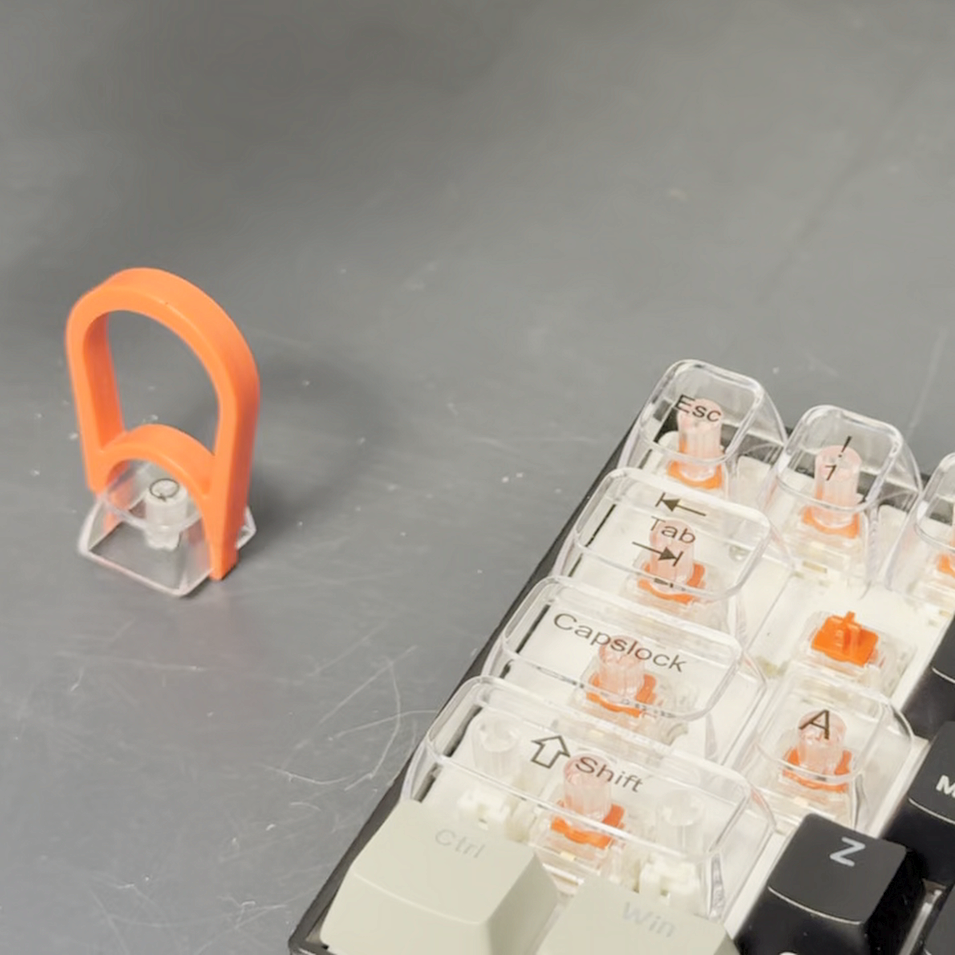}
    \end{minipage}

    \caption{\textbf{Three real-robot tasks.}
    \emph{Top:} \texttt{t-cube}. The robot needs to move two colored
    cubes and carefully balance them on top of the yellow cube.
    \emph{Middle:} \texttt{cable}. The robot needs to feed a USB-C cable
    through a desktop machine case port hole that is barely bigger than
    the cable itself. 
    \emph{Bottom:} \texttt{keycap}. The robot needs to use a plastic tool to extract a transparent keycap from a keyboard.}
    \label{fig:tasks}
\end{figure}

\textbf{Task} \texttt{t-cube}. The scene contains three acrylic cubes (red, pink, and yellow). The robot must grasp the two cubes (red and pink) together and carefully balance it on top of an equal-sized yellow cube. This task is inspired by the T-Block task from \citet{ghugare2025builderbench}. Balancing the two cubes on top of the yellow cube requires a lot of precision. A misalignment of a few millimeters would fail the balance. Nearly all failures of this task occur at placement, making this the most precision-demanding task. An episode succeeds when both cubes rest on top of the yellow cube without falling off. 

\textbf{Task} \texttt{cable}. The scene contains a side panel of a machine case and a USB-C cable. The robot must pick up the cable lying on the table and then insert it through a port hole on the side panel that is barely bigger than the cable head. The insertion is a contact-rich process where the robot must search locally and find the right motion that allows the USB-C to go through the port hole. An episode succeeds when the USB-C passes through the port hole and rests there without falling (with the arm away).

\textbf{Task} \texttt{keycap}. The scene contains a mechanical keyboard and a keycap-puller tool. The robot must pickup the keycap tool and remove a transparent ``Q'' keycap from the keyboard. This requires the arm to seat the tool's two hooks on the opposite sides of the keycap, loosen the keycap, and then pull it off the switch. An episode succeeds when the keycap is detached and placed on the table. 

\section{Full Results}

\begin{table}[H]
  \centering
  \caption{Task success rates (mean $\pm$ standard error). The results for \texttt{\{t-cube/cable/keycap\}} are averaged over \texttt{\{9/5/5\}} seeds with 20 evaluation trials each. Best result per column in bold.}
  \label{tab:success-rates}
  \scalebox{0.87}{
      \begin{tabular}{lcccccc}
        \toprule
        & \multicolumn{2}{c}{\texttt{cable}} & \multicolumn{2}{c}{\texttt{t-cube}} & \multicolumn{2}{c}{\texttt{keycap}} \\
        \cmidrule(lr){2-3} \cmidrule(lr){4-5} \cmidrule(lr){6-7}
         & \textbf{BC} & \textbf{RL} & \textbf{BC} & \textbf{RL} & \textbf{BC} & \textbf{RL} \\
        \midrule
        \emph{Clean} & $0.37 \pm 0.06$ & $0.35 \pm 0.07$ & $0.38 \pm 0.04$ & $0.42 \pm 0.06$ & $\mathbf{0.63 \pm 0.09}$ & $\mathbf{0.81 \pm 0.06}$ \\
        \emph{Random}    & $0.48 \pm 0.09$ & $0.56 \pm 0.04$ & $0.40 \pm 0.05$ & $0.42 \pm 0.03$ & $\mathbf{0.69 \pm 0.06}$ & $0.67 \pm 0.09$ \\
        {\color{dmc}\emph{Ours}}     & {\color{dmc}$\mathbf{0.66 \pm 0.04}$} & {\color{dmc}$\mathbf{0.67 \pm 0.06}$} & {\color{dmc}$\mathbf{0.50 \pm 0.03}$} & {\color{dmc}$\mathbf{0.59 \pm 0.02}$} & {\color{dmc}$\mathbf{0.72 \pm 0.05}$} & {\color{dmc}$\mathbf{0.84 \pm 0.06}$} \\
        \bottomrule
      \end{tabular}
  }
\end{table}

\begin{table}[H]
  \centering
  \caption{Per-seed successes on \texttt{cable} (successes out of 20
    trials per entry). The last row reports the mean success rate
    $\pm$ standard error over the five seeds.}
  \label{tab:cable-per-seed}
  \begin{tabular}{ccccccc}
    \toprule
    & \multicolumn{2}{c}{\emph{Clean}} & \multicolumn{2}{c}{\emph{Random}} & \multicolumn{2}{c}{{\color{dmc}\emph{Ours}}} \\
    \cmidrule(lr){2-3} \cmidrule(lr){4-5} \cmidrule(lr){6-7}
    Seed & \textbf{BC} & \textbf{RL} & \textbf{BC} & \textbf{RL} & \textbf{BC} & \textbf{RL} \\
    \midrule
    0 & $\hphantom{0}4/20$ & $\hphantom{0}2/20$ & $\hphantom{0}6/20$ & $\hphantom{0}9/20$ & $14/20$ & $14/20$ \\
    1 & $11/20$ & $\hphantom{0}9/20$ & $13/20$ & $13/20$ & $12/20$ & $17/20$ \\
    2 & $\hphantom{0}7/20$ & $\hphantom{0}7/20$ & $\hphantom{0}7/20$ & $12/20$ & $12/20$ & $10/20$ \\
    3 & $\hphantom{0}9/20$ & $\hphantom{0}9/20$ & $\hphantom{0}7/20$ & $10/20$ & $12/20$ & $13/20$ \\
    4 & $\hphantom{0}6/20$ & $\hphantom{0}8/20$ & $15/20$ & $12/20$ & $16/20$ & $13/20$ \\
    \midrule
    Mean & $0.37 \pm 0.06$ & $0.35 \pm 0.07$ & $0.48 \pm 0.09$ & $0.56 \pm 0.04$ & {\color{dmc}$\mathbf{0.66 \pm 0.04}$} & {\color{dmc}$\mathbf{0.67 \pm 0.06}$} \\
    \bottomrule
  \end{tabular}
\end{table}

\begin{table}[H]
  \centering
  \caption{Per-seed successes on \texttt{t-cube} (successes out of 20
    trials per entry). The last row reports the mean success rate
    $\pm$ standard error over the nine seeds.}
  \label{tab:cube-per-seed}
  \begin{tabular}{ccccccc}
    \toprule
    & \multicolumn{2}{c}{\emph{Clean}} & \multicolumn{2}{c}{\emph{Random}} & \multicolumn{2}{c}{{\color{dmc}\emph{Ours}}} \\
    \cmidrule(lr){2-3} \cmidrule(lr){4-5} \cmidrule(lr){6-7}
    Seed & \textbf{BC} & \textbf{RL} & \textbf{BC} & \textbf{RL} & \textbf{BC} & \textbf{RL} \\
    \midrule
    0 & $\hphantom{0}9/20$ & $10/20$ & $\hphantom{0}4/20$ & $\hphantom{0}9/20$ & $\hphantom{0}9/20$ & $12/20$ \\
    1 & $\hphantom{0}3/20$ & $\hphantom{0}0/20$ & $\hphantom{0}8/20$ & $\hphantom{0}7/20$ & $\hphantom{0}8/20$ & $\hphantom{0}9/20$ \\
    2 & $10/20$ & $12/20$ & $11/20$ & $12/20$ & $11/20$ & $12/20$ \\
    3 & $11/20$ & $\hphantom{0}9/20$ & $\hphantom{0}6/20$ & $\hphantom{0}7/20$ & $11/20$ & $12/20$ \\
    4 & $\hphantom{0}8/20$ & $\hphantom{0}9/20$ & $\hphantom{0}7/20$ & $\hphantom{0}7/20$ & $10/20$ & $11/20$ \\
    5 & $\hphantom{0}8/20$ & $\hphantom{0}8/20$ & $\hphantom{0}4/20$ & $\hphantom{0}8/20$ & $10/20$ & $12/20$ \\
    6 & $\hphantom{0}6/20$ & $\hphantom{0}9/20$ & $\hphantom{0}9/20$ & $\hphantom{0}8/20$ & $10/20$ & $13/20$ \\
    7 & $\hphantom{0}5/20$ & $\hphantom{0}6/20$ & $14/20$ & $\hphantom{0}9/20$ & $\hphantom{0}8/20$ & $12/20$ \\
    8 & $\hphantom{0}8/20$ & $12/20$ & $\hphantom{0}9/20$ & $\hphantom{0}8/20$ & $13/20$ & $14/20$ \\
    \midrule
    Mean & $0.38 \pm 0.04$ & $0.42 \pm 0.06$ & $0.40 \pm 0.05$ & $0.42 \pm 0.03$ & {\color{dmc}$\mathbf{0.50 \pm 0.03}$} & {\color{dmc}$\mathbf{0.59 \pm 0.02}$} \\
    \bottomrule
  \end{tabular}
\end{table}

\begin{table}[H]
  \centering
  \caption{Per-seed successes on \texttt{keycap} (successes out of 20
    trials per entry). The last row reports the
    mean success rate $\pm$ standard error over the five seeds.}
  \label{tab:keycap-per-seed}
  \begin{tabular}{ccccccc}
    \toprule
    & \multicolumn{2}{c}{\emph{Clean}} & \multicolumn{2}{c}{\emph{Random}} & \multicolumn{2}{c}{{\color{dmc}\emph{Ours}}} \\
    \cmidrule(lr){2-3} \cmidrule(lr){4-5} \cmidrule(lr){6-7}
    Seed & \textbf{BC} & \textbf{RL} & \textbf{BC} & \textbf{RL} & \textbf{BC} & \textbf{RL} \\
    \midrule
    0 & $14/20$ & $19/20$ & $17/20$ & $19/20$ & $17/20$ & $20/20$ \\
    1 & $14/20$ & $15/20$ & $15/20$ & $13/20$ & $17/20$ & $18/20$ \\
    2 & $\hphantom{0}6/20$ & $13/20$ & $15/20$ & $15/20$ & $13/20$ & $17/20$ \\
    3 & $13/20$ & $15/20$ & $11/20$ & $\hphantom{0}9/20$ & $12/20$ & $16/20$ \\
    4 & $16/20$ & $19/20$ & $11/20$ & $11/20$ & $13/20$ & $13/20$ \\
    \midrule
    Mean & $0.63 \pm 0.09$ & $0.81 \pm 0.06$ & $0.69 \pm 0.06$ & $0.67 \pm 0.09$ & {\color{dmc}$\mathbf{0.72 \pm 0.05}$} & {\color{dmc}$\mathbf{0.84 \pm 0.06}$} \\
    \bottomrule
  \end{tabular}
\end{table}

\begin{table}[H]
  \centering
  \caption{Success rates during data collection, computed from the
    recorded episodes. Each entry reports successful episodes over
    all collected episodes and the corresponding percentage.}
  \label{tab:collection-success}
  \begin{tabular}{lccc}
    \toprule
    & \texttt{cable} & \texttt{t-cube} & \texttt{keycap} \\
    \midrule
    \emph{Clean}  & $252/260$ (96.9\%) & $432/749$ (57.7\%) & $113/118$ (95.8\%) \\
    \emph{Random} & $233/244$ (95.5\%) & $311/686$ (45.3\%) & $122/123$ (99.2\%) \\
    {\color{dmc}\emph{Ours}} & {\color{dmc}$178/189$ (94.2\%)} & {\color{dmc}$361/663$ (54.4\%)} & {\color{dmc}$110/117$ (94.0\%)} \\
    \bottomrule
  \end{tabular}
\end{table}

\section{Hyperparameters}
\label{appendix:hyperparameters}
\begin{table}[H]
    \centering
    \begin{tabular}{@{}cc@{}}
        \toprule
        \textbf{Parameter} & \textbf{Value} \\
        \midrule
        Batch size &  $256$ \\
        Discount factor ($\gamma$) & $0.97$ \\
        Action chunk size ($h$) & $5$ \\
        Best-of-N sampling ($N$) & $4$ \\ 
        Optimizer & Adam \\
        Learning rate & $3 \times 10^{-4}$ \\
        Target network update rate ($\lambda$) & $5 \times 10^{-3}$ \\
        Target entropy for residual Gaussian policy & $-5 \times |\mathcal{A}|$ \\
        Zero residual action probability & $0.1$ \\
        Maximum residual action magnitude & $\|\Delta a\|_\infty \leq 0.1$ \\
        Network width & $512$ \\
        Network depth & $4$ hidden layers \\
        Critic pessimistic backup ($\rho$) & $2$ \\
        Critic ensemble size & $10$ \\
        Policy flow steps ($T$) & $10$ \\
        Action delay ($\delta$) & $\SI{0.05}{\second}$ \\
        Control frequency ($F$) & $\SI{10}{\hertz}$ \\
    \bottomrule
    \end{tabular}
    \vspace{1mm}
    \caption{ \textbf{Shared hyperparameters.}}
    \label{tab:shared-hyperparams}
\end{table}

\begin{table}[H]
    \centering
\scalebox{1.0}{
    \begin{tabular}{@{}cccccc@{}}
    \toprule
            Tasks                  & Random Noise & Ours & Training Steps & Data Size \\
   & $c$ & $\beta$  \\
    \midrule
        \texttt{t-cube} & $0.05$ &  $10$  & $1 \times 10^{5}$& $10000$\\
        \texttt{cable} & $0.05$ &  $0.03$ & $0.5 \times 10^{5}$& $4000$\\
        \texttt{keycap} & $0.1$ &  $0.01$ & $0.5 \times 10^{5}$ & $3000$\\
    \bottomrule
    \end{tabular}
    }
    \vspace{1mm}
    \caption{\textbf{Task-specific hyperparameters.}}
    \label{tab:task-hparam}
\end{table}

\end{document}